\pdfoutput=1

\documentclass[11pt]{article}

\usepackage{acl}

\usepackage{times}
\usepackage{latexsym}
\usepackage[T1]{fontenc}

\usepackage[utf8]{inputenc}

\usepackage{microtype}

\usepackage{algorithm}
\usepackage{algpseudocode}
\usepackage{graphicx}
\usepackage{enumitem}

\makeatletter
\def\@fnsymbol#1{\ensuremath{\ifcase#1\or \dagger\or \ddagger\or
   \mathsection\or \mathparagraph\or \|\or **\or \dagger\dagger
   \or \ddagger\ddagger \else\@ctrerr\fi}}
    \makeatother

\title{Targeted Extraction of Temporal Facts from Textual Resources for Improved Temporal Question Answering over Knowledge Bases}

\author{{Nithish Kannen$^1$}\thanks{~~This author's contribution to this work is while he was interning at IBM Research AI, India.}~, Udit Sharma$^2$, Sumit Neelam$^2$, Dinesh Khandelwal$^2$ \\
  \bf Shajith Ikbal$^2$, Hima Karanam$^2$, L Venkata Subramaniam$^2$ \\
  $^1$Indian Institute of Technology, Kharagpur, India \\
  $^2$IBM Research AI, India \\
  \normalsize nithishkannen@gmail.com, 
  \{sumit.neelam, udit.sharma, dikhand1, \\ \normalsize shajmoha, hkaranam, lvsubram\}@in.ibm.com \\
  }

\begin{document}
\maketitle

\newcommand\tab[1][0.4cm]{\hspace*{#1}}

\begin{abstract}
Knowledge Base Question Answering (KBQA) systems have the goal of answering complex natural language questions by reasoning over relevant facts retrieved from Knowledge Bases (KB). One of the major challenges faced by these systems is their inability to retrieve all relevant facts due to factors such as incomplete KB and entity/relation linking errors. In this paper, we address this particular challenge for systems handling a specific category of questions called temporal questions, where answer derivation involve reasoning over facts asserting point/intervals of time for various events. 
We propose a novel approach where a targeted temporal fact extraction technique is used to assist KBQA whenever it fails to retrieve temporal facts from the KB. We use $\lambda$-expressions of the questions to logically represent the component facts and the reasoning steps needed to derive the answer. This allows us to spot those facts that failed to get retrieved from the KB and generate textual queries to extract them from the textual resources in an open-domain question answering fashion.
We evaluated our approach on a benchmark temporal question answering dataset considering Wikidata and Wikipedia respectively as the KB and textual resource. Experimental results show a significant $\sim$30\% relative improvement in answer accuracy, demonstrating the effectiveness of our approach.

\end{abstract}
\section{Introduction}

Complex Question Answering has caught the interest of researchers working in NLP and Semantic Web lately, and is emerging as an important research topic with many potential applications \cite{svitlana2019, maheshwari2018learning, saxena-etal-2020-improving, fu2020survey, bhutani-etal-2020-answering, wu2021modeling, neelam2021}. Answering complex questions involve integration of multiple facts identified and extracted from disjoint pieces of information. Two critical components in systems trying to achieve this are: 1) knowledge source - to retrieve/extract relevant facts, and 2) reasoning - to integrate those facts into final answer. 
Majority of the past work on Question Answering (QA) has focused primarily on one of these components and hence are limited in their ability to answer complex questions.

QA work in NLP community has focused primarily on using textual data as knowledge resource. Text-corpus based QA systems, studied in vast detail over the last few years, have achieved impressive answer accuracies \cite{zhang2020retrospective, zhang2019sgnet, saha-etal-2018-duorc}. However, they are largely limited to simple questions, because reasoning over a large amount of unstructured knowledge in text is difficult with the current text-based techniques. On the other hand, QA work in Semantic Web community has primarily focused on Knowledge Base Question Answering (KBQA) systems, where relevant facts needed to answer the question are retrieved from Knowledge Base (KB) and reasoned over. Although structured knowledge in KB make the KBQA systems better-equipped to handle complex reasoning  \cite{bhutani-etal-2020-answering, svitlana2019, wu2021modeling}, they suffer from issue of incomplete knowledge, resulting from difficulty in gathering and curating large amount of structured knowledge \cite{jain2020temporal, lacroix2020tensor, garcia2018}.

\begin{figure*}[t!]
  \centering
  \includegraphics[keepaspectratio,width=0.90\textwidth]{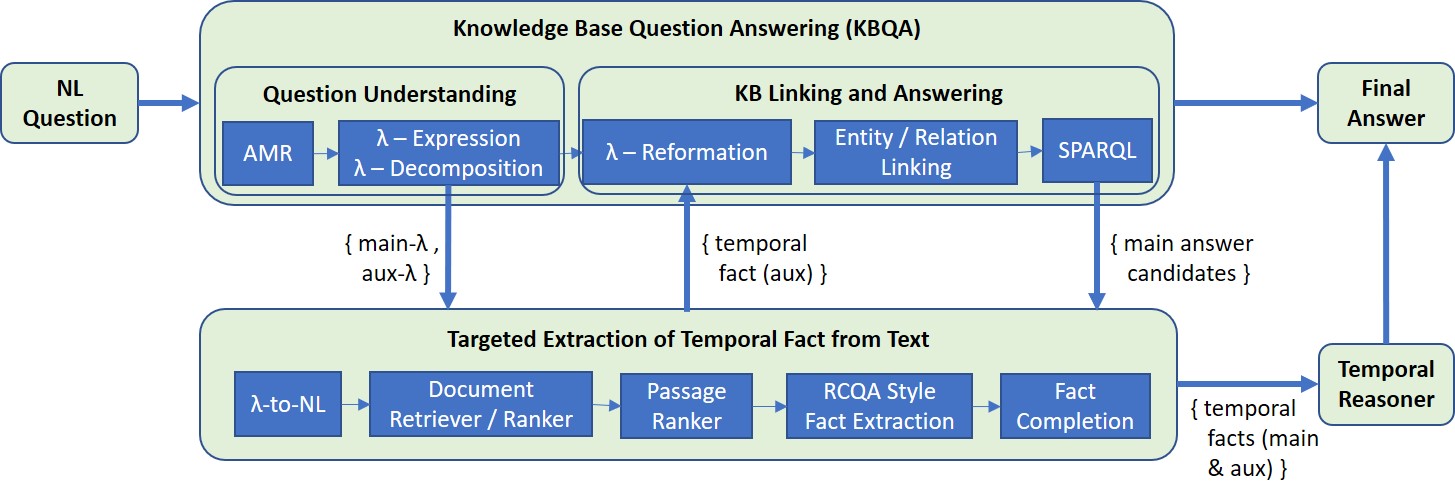}
  \caption{An illustration of the proposed approach. Upper line of modules correspond to the KBQA pipeline, while lower line of modules are related to targeted fact extraction from textual resources. 
  This picture should be seen together with Algorithm \ref{alg:overall}, which gives algorithm for the overall approach.
  }
  \vspace{-0.3cm}
  \label{fig:overall}
  \vspace{-0.3cm}
\end{figure*}

In this paper, we present a novel combination of successful elements from the past approaches, i.e., KB based and text-corpus based, towards the goal of answering a specific category of complex questions called, temporal questions.
Answering temporal questions additionally involve reasoning over temporal facts, i.e., assertions on points and intervals in time of events\footnote{In this paper, we consider entities and facts (represented as triples in KBs) with associated time intervals as events. For example, \{\textit{World War 2}, (start time:1939, end time:1945)\} and \{(\textit{Franklin D. Roosevelt}, \textit{President of}, \textit{United States}),  (start time:1933, end time: 1945)\} are events.}.
For example, to answer question \textit{Who was the President of the United States during World War 2?} (in illustration Figure \ref{fig:lambda}), we need to know the time period of an event named \textit{World War 2} and the list of all US presidents and their corresponding time intervals in office, so we can find those whose time intervals in office overlap with that of the \textit{World War 2}.

In our approach, we use a targeted extraction technique to assist KBQA at points where it fails. KBQA fails whenever it fails to retrieve all the relevant facts needed to answer the question\footnote{Note that this issue is even more pronounced in case of temporal facts, because adding temporal facts to the KBs has started gaining momentum only in recent times.}, because of reasons such as \textit{incomplete KB} and \textit{inaccurate entity/relation linking}. The goal of targeted extraction is to look for and extract from textual resources those facts that failed to get served by the KB. Upon successful extraction from text, the extracted facts can be used together with the facts successfully retrieved from the KB to derive the final answer.
To facilitate identification of KB failures and subsequent extractions, we use a semantic parsing approach that transforms natural language (NL) question into $\lambda$-expression \cite{neelam2021}.
$\lambda$-expression of a question logically represents the component facts needed from the knowledge source and the reasoning needed to be performed over those facts to derive the final answer. 

We limit the scope of our targeted extraction to temporal facts. This is because, extraction in general from textual resources is known to be noisy. Our hope is that a restricted extraction approach (i.e., targeted extraction of simple temporal facts) would likely result in relatively more accurate extractions. This way our approach tries to make an effective utilization of the KB (reliable but not exhaustive) and the textual resources (vast but noisy).
Targeted temporal fact extraction technique in our approach is devised as open-domain question answering system to answer simple temporal factoid questions. For each temporal fact that failed to get served by the KB, we generate equivalent NL question from the corresponding component $\lambda$-expression, to search for that fact from textual resources and fetch if available. For example, if KB fails to serve temporal fact related to \textit{Word War 2}, we form a NL question \textit{When was World war 2?} Getting an answer for this question in a open-domain question answering fashion is equivalently extracting temporal fact related to \textit{World war 2}. 
In the work for this paper, we use Wikidata as the KB and Wikipedia as the textual resource. The base KBQA system used in our approach is built  in a modular fashion similar to \cite{neelam2021}.

The main contributions of our work are:

\noindent
1. We propose a novel combination of systems performing \textit{KBQA} and \textit{Targeted extraction from text-corpus} for improved temporal question answering, where \textit{Targeted extraction} is used only to assist KBQA wherever it fails. 

\noindent
2. A $\lambda$-calculus based semantic representation and decomposition of the question to help find KBQA gaps. 
    $\lambda$-expressions of the question logically represents set of facts needed from the KB. Those facts that fail to get fetched from the KB correspond to the KBQA gaps.

\noindent
3. An (open-domain question answering style) approach for targeted extraction of temporal facts from textual resources, to cover KBQA gaps. Those components of $\lambda$-expression that fail to fetch facts from the KB are converted into NL queries, to first retrieve relevant text snippets and then extract the answer.

\noindent
4. Experimental evaluation demonstrating effectiveness of the proposed approach showing $\sim$30\% relative improvement in answer accuracy metric.

\vspace{-0.2cm}
\section{Related Work}

Although Complex KBQA has been an active research topic \cite{svitlana2019, saxena-etal-2020-improving, wu2021modeling, shi2020kqa}, there has been very limited research focused on Temporal KBQA. 
Temporal Questions require identification of time intervals of events and temporal reasoning.

\subsection{Temporal KBQA Datasets:}
TempQuestions \cite{Jia2018a} is one of the first publicly available temporal KBQA dataset consisting of 1271 questions. 
However, this dataset was annotated over FreeBase, which is no longer maintained and was officially discontinued in 2014. SYGMA~\cite{neelam2021} introduced a subset of TempQuestions that can be answered over wikidata called TempQA-WD. We use both TempQA-WD and full TempQuestions data sets to evaluate our approach.
CRONQUESTION~\cite{saxena-etal-2021-question} is another temporal KBQA dataset that uses its own KB drawn from Wikidata. 
Event-QA dataset~\cite{eventkg} is based on Event-KG, curated from DBpedia, Wikidata and YAGO.
Since these datasets are generated in a template based manner using existing facts from the KBs, they do not represent the real world challenge of incomplete KBs. One of the main goals of our approach is to handle the issue of incomplete KBs. 

\subsection{Temporal KBQA Systems:} TEQUILA \cite{Jia2018b} is one of the first 
attempts to address temporal question answering over KBs. 
It used an existing KBQA engine \cite{Abujabal2017} to answer individual sub-questions and perform a temporal reasoning over the answers to derive the final answer.
SYGMA~\cite{neelam2021} is another system that works on a Wikidata and uses $\lambda$-expressions to represent the facts and their temporal reasoning operators. 
TEQUILA uses a pre-specified set of temporal signals (10 signal words) to decompose questions into sub-questions at sentence level in a rule-based manner. Instead, we follow the approach similar to SYGMA that use a sophisticated semantic parsing approach involving AMR (Abstract Meaning Representation) \cite{banarescu2013} and $\lambda$-calculus \cite{zettlemoyer2012learning} to get logical representations of the questions. 
This enables decomposition of the questions at semantic level and is likely robust to linguistic variations as well. 

\subsection{{KB + Text for QA}}

There have been past work exploring effectiveness of using KB and text resources for complex QA \cite{sun2018open, xiong2019improving}. However, none of them address the temporal context addressed in this work.
Prior work using a combination of KB and text  have largely been based on end-to-end neural models. GRAFT-Net \cite{sun2018open} constructs a sub-graph from KB and text corpora using an early fusion technique. The task of QA is then reduced to binary classification over the nodes of this sub-graph. PullNet \cite{sun2019pullnet} proposes to build sub-graph through an iterative process\cite{xiong2019improving}, utilise a graph-attention based KB reader and knowledge-aware text reader.

All these methods are based on end-to-end neural models that require large amount of training data and offer little interpretability, which is essential to evaluate 
intermediate stages of complex QA systems. 
Additionally, labeling large amounts of data for KBQA is hard 
\cite{trivedi2017lc}.
In this work, we extend modular approach described in ~\cite{neelam2021}, additionally incorporating it with a targeted extraction pipeline. We made this choice as this particular approach integrates multiple, reusable modules that are pre-trained for their specific individual tasks (semantic parsers, entity and relational linkers, rankers  and re-rankers and reading comprehension model) thus offering interpretability and flexibility for optimal combination of textual extraction with KBQA. Additionally, this
does not require a large amount of domain-specific training data. 

\subsection{Question Decomposition}
Our work uses a form of logical query decomposition, based on $\lambda$-expression of the NL question,
to help effectively combine the KB with the text resources.
Some of the past work in the literature on QA have also explored question decomposition.
BREAK IT down~\cite{Wolfson2020Break} is a popular benchmark data that captures complex question as an ordered list of tasks, that when executed in sequence will derive the final answer. It introduced question decomposition meaning representation (QDMR) to represent decomposed questions in an intermediate form resembling SQL. TEQUILA~\cite{Jia2018b} used temporal signal (words) based question decomposition to turn natural language questions into sub questions.

~
\vspace{-0.5cm}
\section{Our Approach}

Figure 1 
shows a block diagram of our proposed approach. 
It consists of two groups of modules:
\vspace{-0.2cm}
\begin{enumerate}[leftmargin=*]
    \item Upper line of modules that are built to get answer purely from the KB, called \textit{KBQA pipeline}.
    \item Lower line of modules to perform targeted temporal fact extraction from the textual resources, called \textit{Extraction pipeline}.
\end{enumerate}
\vspace{-0.2cm}
The overall strategy, in our approach to derive answer, is to rely on \textit{KBQA pipeline} to the maximum possible extent (because of reliability of information in KB) and use \textit{Extraction pipeline} only to aid KBQA wherever it fails.
Accordingly, given a question, first we try to get an answer purely from the \textit{KBQA pipeline}. If that fails, we investigate the reasons for the failure of \textit{KBQA pipeline} and make a targeted use of the \textit{Extraction pipeline} to compensate for those failures.
Algorithm \ref{alg:overall} (described in detail later in the paper) gives the flow sequence among various modules within the block diagram, starting from question until reaching the final answer.




\textit{KBQA pipeline} in our approach is built in a modular fashion similar to \cite{neelam2021, pavan2020}. Various modules in it transform the question step-by-step along the pipeline through intermediate representations to finally represent the question in terms of the KB elements and then to compute the final answer.
This is achieved in two steps: 

\noindent
1. \textit{Question Understanding} - to transform questions into their logical representations and decompose them as per the mentions of events in them and the associated temporal fact requirements. We use $\lambda$-calculus for logical representation, i.e., questions are transformed into $\lambda$-expressions, that: a) compactly specifies the set of facts needed from the KB and the reasoning needed to be performed, and b) offers flexibility to perform event-based decomposition.

\noindent
2. \textit{KB Linking and Answering} - to map the elements of the $\lambda$-expression onto the KB elements, so that the corresponding fact fetching could be executed on the KB.

In our approach, the use of $\lambda$-expression plays a critical role in locating the points of failures of the \textit{KBQA pipeline} and performing further targeted search/extraction using \textit{Extraction pipeline}.
We give a detailed description of our approach in the following subsections.
Next two sub-sections give a description of \textit{Question Understanding} and \textit{KB Linking and Answering} in \textit{KBQA pipeline}. Then we describe how \textit{Extraction pipeline} is used to handle \textit{KBQA pipeline} failures, followed by a description of the \textit{Extraction pipeline} itself.

%

\subsection{Question Understanding}
\label{sec:qu}

\begin{figure}[!ht]
  \centering
   \includegraphics[keepaspectratio, width=0.45\textwidth]{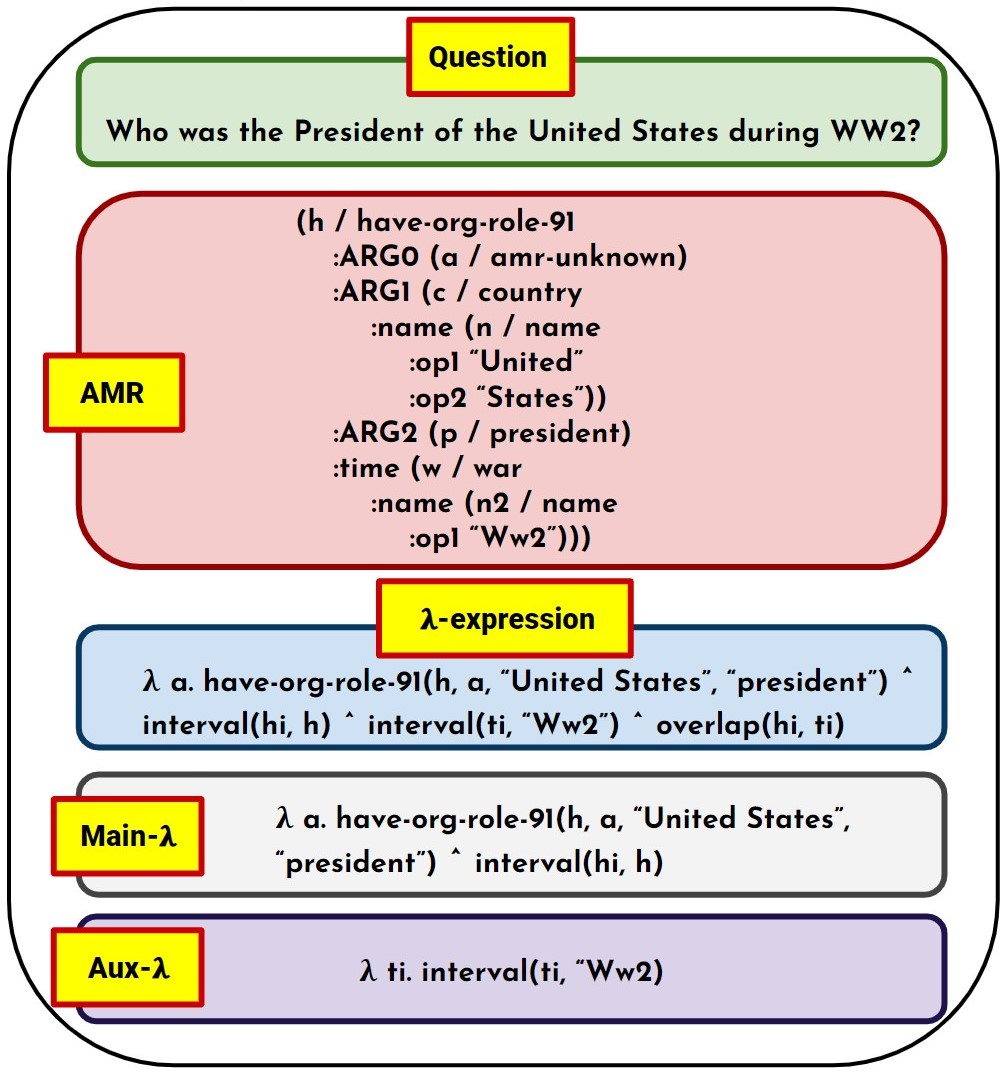}
  \caption{Question understanding. Example question, its AMR, $\lambda$-expression and its decomposition. }
  \label{fig:lambda} 
  \vspace{-0.5cm}
\end{figure}

The goal of \textit{Question Understanding} is to 1) transform NL questions into corresponding $\lambda$-expressions that logically represent the set of event specific facts needed from the KB and the reasoning needed to be performed to derive the answer and 2) further perform event-specific decomposition.
We use method as in \cite{neelam2021} to construct $\lambda$-expressions of the questions from their AMR (Abstract Meaning Representation) \cite{banarescu2013}. 
AMR encodes meaning of the sentence into a rooted directed acyclic graph where nodes and edges represent concepts and relations respectively. Such a representation is useful because event-specific decomposition of the question is represented to some extent as the sub-paths and sub-graphs in the AMR graph.
Figure \ref{fig:lambda} shows an illustration of AMR and $\lambda$-expression for our example question.
This example illustrates how $\lambda$-expression compactly represents, the mentions of events in the question (as its sub-components), facts about those events (that need to be fetched from the knowledge source), and the reasoning steps (that need to be performed to derive the final answer).

\if false
\begin{algorithm}
\caption{AMR to $\lambda$-expression}\label{alg:amr2lambda}
\begin{algorithmic}
\Procedure{Convert}{$amr\_root$}:
\State $expr = empty\_expr()$
\If{$multi\_event(amr\_root)$}
\State $amr\_list$ = $break\_amr(amr\_root)$
\State $connective$ = 
\State $~~~~~get\_temp\_connective(amr\_root, amr\_list)$
\For {each $amr$ in $amr\_list$}:
\State $sub\_expr = Convert(amr)$
\State append $sub\_expr$ to $expr$
\EndFor
\State append $connective$ to $expr$
\Else
\For{each $child$ in $children(amr\_root)$}:
\State $child\_expr = Convert(child)$
\State append $child\_expr$ to $expr$
\EndFor
\State $expr = get\_expr(amr\_root, expr)$
\EndIf
\State \textbf{return} $expr$
\end{algorithmic}
\end{algorithm}
\fi

$\lambda$-expression constructed for the question is further processed to decompose into components: 

\vspace{0.1cm}
\noindent
1. \textit{Main-$\lambda$}: Part of the $\lambda$-expression related to the unknown variable, i.e., main event being questioned. For example in Figure \ref{fig:lambda}, $a$ is the unknown variable, whose value if found is answer to the question. 

\noindent
2. \textit{Aux-$\lambda$}: part of the $\lambda$-expression not related to the unknown variable, but related to the rest of the events mentioned in the question. This part serves the purpose of adding temporal constraint to the candidate answer values for the unknown variable. 

We use a rule based approach to perform decomposition, that simply uses unknown variable as anchor to segregate the respective components.
Note that the decomposed $\lambda$-expressions play a critical role in our approach to identify the points of failures of the \textit{KBQA pipeline} and to further decide on the use of \textit{Extraction pipeline}.
Figure \ref{fig:lambda} shows main-$\lambda$ and aux-$\lambda$ for our example question.

\subsection{KB Linking and Answering}

This is essentially a step to ground the Entity and Relation mentions of $\lambda$-expression to the KB, i.e., map the elements of $\lambda$-expression onto the corresponding KB elements, so that fact fetching can be executed on KB.
Relation mentions are the predicates (for example, \textit{have-org-role} in Figure \ref{fig:lambda}) and entity mentions are the arguments (for example, \textit{United States}, \textit{president}, and \textit{Ww2} in Figure \ref{fig:lambda}). The goal of linking is to map, for example \textit{Ww2} to a node in KB corresponding to \textit{World War 2} (Wikidata id wd:Q362).
After linking, we generate corresponding SPARQL queries that when executed on the KB endpoint would fetch the intended KB facts.
Our approach to linking and SPARQL generation are similar to that of \cite{neelam2021}.

\subsection{Targeted Temporal Fact Extraction from Text When KBQA Fails}
As described earlier, 
our goal is to use textual resources wherever KBQA fails.
KBQA failure can happen because of two reasons: 

\noindent
1. \textit{Linking failure} - when KB linking step fails to successfully map mentions in the $\lambda$-expression to the corresponding KB entities and relations. For example, in Figure \ref{fig:lambda} when mention \textit{Ww2} fails to get linked to the \textit{World War 2} node in the KB.

\noindent
2. \textit{Missing facts} - KBs are known to be incomplete, and hence may fail to fetch a specific fact, simply because it is not present in the KB. For example, if temporal information corresponding to \textit{World War 2} is not present in the KB, trial to fetch time interval corresponding to $\lambda$-expression part interval($ti$, “Ww2") would fail.

$\lambda$-expression specifies all facts that need to be fetched from the KB.
A failure to fetch even a single fact would block KBQA from computing the final answer. 
However, to handle such failures we need to know the specific facts that failed to get fetched from the KB, so that we can look for them into the textual resources.
For this purpose, we categorize KBQA failure as below based decomposed $\lambda$-expressions where failure happens:

\noindent
1. \textit{Aux Failure} - for example, in Figure \ref{fig:lambda}, when time interval of \textit{World War 2} is missing in KB.

\noindent
2. \textit{Main Failure} - for example, in Figure \ref{fig:lambda}, when time intervals of \textit{Franklin D. Roosevelt} and \textit{Harry S. Truman} (who were the US presidents during World War 2) as presidents in the office are missing in the KB.

\begin{algorithm}[ht]
\caption{Algorithm for the overall approach illustrated in Figure 1 
and its flow sequence.}\label{alg:overall}
\small
\begin{algorithmic}
\State $lambda=$   $GetLambda(question)$
\State $ans\_list=$   $GetKBAnswer(lambda)$
\If{$ans\_list$ is empty} \Comment{KBQA Failure}
    \State $ main, aux = Decompose\_Lambda(lambda)$
    \State $ans\_list = GetKBAnswer(aux)$  
    
    \If{$ans\_list$ is empty}  \Comment{Aux Failure}
        \State $fact = Extract\_From\_Text(aux)$
        \State $aux\_fact = fact$  \Comment{for later use}
        \State $reformed\_lambda=$
        \State $~~~~~~~~~ReformLambda(fact, lambda)$
        \State $ans\_list=$
        \State $~~~~~~~~~GetKBAnswer(reformed\_lambda)$
        \If{$ans\_list$ is not empty}
            \State return  \Comment{Ans Found}
            
        \EndIf  
    \EndIf
    
    \State $ans\_list=$  $GetKBAnswer(main)$
    \State $candidate\_facts = []$
    \For {$candidate$ in $ans\_list$}
        \State $fact = Extract\_From\_Text(candidate)$
        \State $candidate\_facts.append({fact, candidate})$
    \EndFor
    
    \State $ans=$
    \State $~~~~~TemporalReasoner(candidate\_facts, auxfact)$
    \State return \Comment{Ans Found}

\Else
 \State return       \Comment{Ans Found, No missing fact in KB}

\EndIf

\end{algorithmic}
\end{algorithm}

In our approach, such a localisation of failure helps in deciding how and when we should use textual resources to compensate for the KBQA failures.
Algorithm \ref{alg:overall} gives our overall approach. 
First we try to get the answer purely from the KBQA pipeline.
If KBQA fails, then we move ahead to evaluate just the Aux-$\lambda$ in the KBQA pipeline. This is expected to result in a time interval, i.e., composed time intervals of all the events that are part of aux-$\lambda$ put together. Failure to get time interval denote that the temporal facts corresponding to events in aux-$\lambda$ are not present in the KB. 
In that case, we use \textit{Extraction pipeline} (which will be described in detail in the next section) to try extract that from the textual resource. Upon successful extraction of time interval for aux-$\lambda$, we construct a reformed $\lambda$-expression from the original $\lambda$-expression by simply replacing auxiliary part with the time interval of aux-$\lambda$. For example, $\lambda$-expression in Figure \ref{fig:lambda} is reformed as:
\vspace{-0.3cm}
\begin{description}
    \item[] $\lambda$ $a$. have-org-role-91($h$, $a$, "United States", "president") $\land$ interval($hi$, $h$) $\land$ overlap($hi$, (interval\_start:1939-09-01, interval\_end:1945-09-02)).
\end{description}
\vspace{-0.3cm}
Note that this reformed $\lambda$-expression do not have aux-$\lambda$. Its NL equivalent is \textit{Who was the President of the United States during period from 1st September 1939 to 2nd September 1945?}
Thus if reformed $\lambda$-expression is passed onto the KBQA pipeline (instead of original $\lambda$-expression), it should result in the same answer, but without fetching facts related to aux-$\lambda$ from the KB.

If \textit{KBQA pipeline} still fails to give answer using reformed $\lambda$-expression, then the issue is because of failure to fetch facts related to main-$\lambda$ from the KB. Hence those facts should be extracted from textual resources using Extraction pipeline. However, we do not fully rely on textual resources for main-$\lambda$, because it represents event with unknown variable. For example, in Figure \ref{fig:lambda} main-$\lambda$ corresponds to the list of all the US presidents and their time intervals in office as the president. It is hard to extract all those information fully from the text. So we make an assumption that we can always get the list of all answer candidates from the KB itself, but may need to look into textual resources only for temporal fact about them. For example, in Figure \ref{fig:lambda} we take that part of the main $\lambda$-expression that would fetch the answer candidates from the KB (leaving out the temporal fact specific components), i.e., \vspace{-0.3cm}
\begin{description}
    \item[] $\lambda$~$a$. have-org-role-91($h$, $a$, "United States", "president")
\end{description}
\vspace{-0.3cm}
and pass that onto the \textit{KBQA pipeline}. Then for each answer candidate obtained we try to extract time interval from the textual resource. For example, if \textit{Franklin D. Roosenvelt} is one of the answer candidates, we try to extract the time interval of \textit{Franklin D. Roosenvelt} being the US president from the textual resources.

Note that we resort to extraction from textual resources only for those facts that failed to get fetched from the KB.
We believe, this approach of targeted extraction from the text (for example to  extract specifically the time period of \textit{Ww2}), is likely to be more accurate than unrestricted extraction, because we are looking to extract facts with a set of known variables and only one unknown variable. 
In fact, this kind of extraction requirement resembles factoid question answering, where text-corpus based QA approaches are known to perform very well. 

\subsubsection{Temporal Reasoning}

Upon successful gathering of all the temporal facts (i.e., time intervals), either from the KB or from the text, we use \textit{Temporal Reasoning} module to chose the final answer from among the answer candidates of the main-$\lambda$. For example, as in Figure \ref{fig:lambda}, $\lambda$-expression component overlap($hi$, $ti$) correspond to the temporal reasoning step, which essentially tries to find the overlap between the time intervals $ti$ (of \textit{Ww2}) and $hi$ (of answer candidates of the main-$\lambda$). This will turn success when those two time intervals overlap and the corresponding answer candidates will be chosen as the final answer. Apart from overlap, we handle other temporal categories such as before, after, now, etc. 
Next we describe \textit{Extraction pipeline}.

\subsection{Extraction pipeline}

In our approach, there are $3$ scenarios where we look to extract facts from textual resources: 1) KBQA fails for aux-$\lambda$ because of Linking failure, 2) KBQA fails for aux-$\lambda$ because of missing temporal fact, and 3) KBQA fails for main-$\lambda$ because of missing temporal fact.
Note that in all these cases, the requirement is to extract temporal facts, either from aux-$\lambda$ or from main-$\lambda$. Our extraction approach as illustrated in Figure 1 (where lower pipeline of modules constitute the \textit{Extraction pipeline}) commonly serves all these requirements.
Our extraction approach resembles open-domain question answering, where first a NL query for missing fact is generated, followed by a retrieval of relevant passages and then a RCQA style answer extraction from those passages.
Next we describe specific modules of the \textit{Extraction pipeline}.



\subsubsection{$\lambda$ to NL Text Query:}
\label{sec:nlquery}
We use simple rules to convert $\lambda$ to NL text query. Since we deal with temporal facts, all the queries start with \textit{When}. Then in case of aux-$\lambda$ we add \textit{was} or \textit{did} depending upon whether the event being considered is entity-based or triple-based.
For example, for the aux-$\lambda$ in Figure \ref{fig:lambda} query is \textit{When was Ww2?}. For triple based events in aux-$\lambda$ example: \{ $\lambda$ $ri$. release-01($r$, "Titanic") $\land$ interval ($ri$, $r$) \}, query is \textit{When did Titanic release?}.
A similar approach is used for main-$\lambda$ answer candidates. For example, for main-$\lambda$ in Figure \ref{fig:lambda} for answer candidate \textit{Franklin D. Roosevelt}, the NL query is \textit{When was Franklin D. Roosevelt president of United States?}.


\subsubsection{Document Retrieval and Passage Ranking}

Once the NL queries are generated, document retrieval is performed in 2 stages. 
First, entities of the question are extracted using BLINK \cite{wu2019general} and Wikipedia pages of all the entities are collected, regardless of whether the extraction is for main-$\lambda$ or aux-$\lambda$.
For example, two entities in Figure \ref{fig:lambda} are \textit{President of United States} and \textit{Ww2}. fuzzywuzzy\footnote{https://pypi.org/project/fuzzywuzzy/} is used to find top matching Wikipedia pages.
Wikipedia content is accessed through Wikipedia API\footnote{https://pypi.org/project/wikipedia/}.
Second, we use NL text query generated from $\lambda$-expression to search documents using MediaWiki API\footnote{https://www.mediawiki.org/wiki/Download}.
We use the Bi-encoder, Cross-Encoder based Siamese-BERT networks \cite{reimers2019sentencebert} to rank passages within the retrieved documents, using NL query of the $\lambda$-expression. Bi-Encoder picks out top 50 relevant passages based on passage-query similarity and Cross-Encoder re-ranks them. Both the models are trained on the MS-MARCO dataset \cite{bajaj2018ms} and we use publicly available pre-trained models\footnote{https://www.sbert.net/}.



\subsubsection{RCQA Style Fact Extraction and Fact Completion}
We use publicly available BERT model\footnote{https://huggingface.co/} (pre-trained on the SQUAD data set \cite{rajpurkar-etal-2016-squad}) to extract fact in a Reading Comprehension QA (RCQA) style, by using NL text query from $\lambda$-expression as the question and top $3$ ranked passages as the context. Interestingly, this is able to achieve good performance without task-specific adaptation.
Note that RCQA style extraction gives one answer, which is sufficient if we are extracting point-in-time. However, for time intervals involving start and end times this will give only the start time or the end time. For this reason, we further take the sentence from which the answer is extracted and use its AMR tree to complete the time interval extraction, by looking at that are siblings to the node with the answer. 

\vspace{-0.2cm}
\section{Experimental Setup}
\label{sec:setup}

\textbf{Data:} 
In our system, we use Wikidata as KB and Wikipedia as textual resource. In our experiments, we evaluated our approach on two aspects: 1) Temporal QA performance of the overall system and 2) Targeted temporal fact extraction performance. For QA performance we used two datasets: 1) \textit{TempQA-WD} \cite{neelam2021} and 2) \textit{TempQuestions} \cite{Jia2018a}.
\textit{TempQA-WD} has $839$ temporal questions and their corresponding Wikidata answers. \textit{TempQuestions} has 1271 temporal questions and their corresponding Freebase answers.
In fact, questions in \textit{TempQA-WD} are a subset of that in \textit{TempQuestions}, but has answers in Wikidata, matching the KB of our system.
We chose to use Wikidata as the KB for our system because:
1) Wikidata is well structured, fast evolving, most up-to-date, actively maintained, and
2) Wikidata supports reification of statements (triples), supporting representation of temporal information such as start time, end time and point in time of events.
These make Wikidata a preferred choice, because through its design it can support complex reasoning requirements and also may likely provide more accurate answers. 
Evaluation of our system with \textit{TempQuestions} (in spite of its Freebase answers) is mainly for comparison to Tequila system. However, the challenge is in matching the gold Freebase answers with the system generated Wikidata answers\footnote{For example, for question \textit{When did the Colts last win the Superbowl?}, Freebase answer is \textit{Super Bowl XLI} while Wikidata answer is \textit{2007}.}.
In spite of this, we decided not to support Freebase in our system because it is officially discontinued in 2014 \cite{freebase}. 
For evaluation of targeted extraction we took a subset of NL queries generated in our system, as described in Section \ref{sec:nlquery}, for which we could get gold temporal facts from KB. We used $3709$ NL queries for this evaluation, collected from the 175 questions of \textit{TempQA-WD} dataset.



\noindent
\textbf{Baseline and Metrics:}
Note that the key aspect of our approach (referred as \textit{KB+Text}) is the support that Extraction pipeline gives to the KBQA, wherever it fails.
To evaluate its effectiveness we considered following baselines:
1) \textit{Only-KB} - when Extraction pipeline is not used to support KBQA, 2) \textit{KB+TemporalText} - when in our system temporal facts are always extracted from the textual resources, and 3) \textit{Open-Domain-QA} - a system that purely works on text, not KB at all.
For \textit{Open-Domain-QA} we used RAG~\cite{RAG}, a state-of-the-art open domain question answering system.
Given a question, RAG first retrieves a set of passages from the text corpus and then uses BART~\cite{lewis2020bart}, a state-of-the-art seq2seq model, to generate the answer given the passages and the question.
We have experimented with both
RAG-Token\footnote{{https://huggingface.co/facebook/rag-token-nq}}
and 
RAG-Sequence\footnote{https://huggingface.co/facebook/rag-sequence-nq}
pretrained models trained on \textit{wiki\_dpr}\footnote{https://huggingface.co/datasets/wiki\_dpr} 
dataset. We report the results of RAG-Token as it performed better. We use standard performance metrics typically used by KBQA systems to report our results, namely macro precision, macro recall and macro F1.
Note that we also use approximate match in addition to the exact match to find overlap, to handle cases of mismatches between system generated answers and gold answers.
\vspace{-0.2cm}
\section{Results and Discussion}
Table \ref{tab:results} shows performance comparison of the proposed and the baseline systems on \textit{TempQA-WD} dataset.
The proposed approach (\textit{KB+Text}) is able to achieve an improvement of 0.095 in F1 score ($\sim$30\% relative improvement) over \textit{Only-KB} baseline, demonstrating the effectiveness of our approach in making a targeted utilization of the textual resources towards assisting KBQA. 
This point is further emphasized by performance comparison of our approach to all other baselines.
Comparison to \textit{KB+TemporalText} illustrates the reliability of facts obtained from KB, whenever available. Performance of RAG is inferior to \textit{Only-KB} illustrating the reasoning capability of a KBQA system in comparison to state-of-the-art text-corpus based system, when faced with complex questions.


\begin{table}[htb]
\begin{small}
\centering
\setlength\tabcolsep{6pt}
\renewcommand{\arraystretch}{1.0}
\begin{tabular}{lccc}
\hline
System & Precision & Recall & F1 \\
\hline
Only-KB  & 0.320 & 0.329 & 0.321 \\
KB+TemporalText & 0.224 & 0.227 & 0.212 \\
Open-Domain QA (RAG) & 0.291 & 0.240 & 0.252 \\
\hline
KB+Text (proposed) & 0.423 & 0.434 & 0.416 \\
\hline
\end{tabular}
\vspace{-0.1cm}
\caption{Performance comparison on \textit{TempQA-WD}.}
\label{tab:results}
\vspace{-0.5cm}
\end{small}
\end{table}

Table \ref{tab:rag_results} shows performance comparison on \textit{TempQuestions} dataset.
We did not evaluate \textit{KB+TemporalText} on this dataset. Instead, in the table, we added the accuracy of Tequila system on the dataset, as found in \cite{Jia2018b}. Note that our approach helped improving accuracy in this dataset too, in comparison to both \textit{Only-KB} and RAG. However, accuracy of our system is well below that of Tequila. The reason is Tequila system is built on Freebase, which is also KB used in \textit{TempQuestions}. 
In contrast, our system is built on Wikidata, thus causing mismatch between the system generated and gold answers. We tried bridging this, but there are many issues that we could not handle (as described in Section \ref{sec:setup}). In fact, this point is highlighted also by the accuracy of \textit{Only-KB} on this dataset, where our manual examination showed many cases of lexical mismatch in spite of being correct answer.
Moreover, Table \ref{tab:results} showed that our system (KB+Text) achieved F1 score of $0.416$ on 839 questions with Wikidata, which is higher than Tequila's $0.367$, although on a subset.


\vspace{-0.2cm}
\begin{table}[htb]
\begin{small}
\centering
\setlength\tabcolsep{6pt}
\renewcommand{\arraystretch}{1.0}
\begin{tabular}{lccc}
\hline
System & Precision & Recall & F1 \\
\hline
Only-KB (baseline) & 0.127 & 0.138 & 0.128 \\
Open-Domain QA (RAG) & 0.241 & 0.207 & 0.215 \\
Tequila \cite{Jia2018b} & 0.360 & 0.423 & 0.367 \\
\hline
KB+Text & 0.269 & 0.277 & 0.260 \\
\hline
\end{tabular}
\vspace{-0.1cm}
\caption{Performance comparison on \textit{TempQuestions}.} 
\label{tab:rag_results}
\vspace{-0.3cm}
\end{small}
\end{table}

Since Extraction pipeline is a critical component in our system, we also evaluated its independent accuracy using a small set of NL queries generated from our system (as described in Section \ref{sec:setup}). Table \ref{tab:res2} show the results. This shows that improvements to the \textit{Extraction pipeline} can help further improve the overall performance of our system.
\vspace{-0.3cm}
\begin{table}[htb]
\begin{small}
\centering
\setlength\tabcolsep{4pt}
\renewcommand{\arraystretch}{1.0}
\begin{tabular}{lccc}
\hline
 & Precision & Recall & F1 \\
\hline
on $3709$ NL queries & 0.171 & 0.163 & 0.165 \\
\hline
\end{tabular}
\vspace{-0.1cm}
\caption{Evaluation of \textit{Extraction pipeline}.}
\label{tab:res2}
\vspace{-0.5cm}
\end{small}
\end{table}

\if false
\vspace{-0.1cm}

\begin{table}[htb]
\begin{small}
\centering
\setlength\tabcolsep{4pt}
\renewcommand{\arraystretch}{1.0}
\begin{tabular}{lccc}
\hline
 & Precision & Recall & F1 \\
\hline
for $442$ where Only-KB failed & 0.274 & 0.278 & 0.258 \\
\hline
\end{tabular}
\vspace{-0.1cm}
\caption{Performance on questions where KBQA has failed.}
\label{tab:res2}
\end{small}
\end{table}
\fi

\if false
\vspace{-0.1cm}
We also evaluated our system on the original TempQuestions dataset of $1271$ questions. Table \ref{tab:rag_results} gives results. 
Note that in this dataset also our approach 
helped improving the overall performance. We have also compared with the performance of the Tequila system (as reported in \cite{Jia2018b}). The accuracy of proposed system is well below that of Tequila. The reason is Tequila is built with Freebase and hence it matches with the annotations of the original TempQuestions dataset, also in Freebase. On the other hand, our system is built on Wikidata and hence there is a mismatch between the answers generated by our system and that of the gold answers from Freebase. We tried bridging that mismatch. But still there are many issues that we could not handle as described already in the \textit{Data} subsection. Our choice of sticking to Wikidata is based on the reasons described in \textit{Data} subsection. Nevertheless, in Table 1 our system (KB+Text) achieves F1 score of $0.414$ on 839 questions that has Wikidata answers, which is higher than Tequila's $0.367$, although it is on a subset.

\begin{table}[htb]
\begin{small}
\centering
\setlength\tabcolsep{6pt}
\renewcommand{\arraystretch}{1.0}
\begin{tabular}{lccc}
\hline
System & Precision & Recall & F1 \\
\hline
Only-KB (baseline) & 0.127 & 0.138 & 0.128 \\
KB+Text (proposed) & 0.269 & 0.277 & 0.260 \\
\hline
Tequila \cite{Jia2018b} & 0.360 & 0.423 & 0.367 \\
\hline
\end{tabular}
\vspace{-0.1cm}
\caption{Performance on original TempQuestions dataset.} 
\label{tab:rag_results}
\end{small}
\end{table}
\fi

\if false
\begin{table}[htb]
\begin{small}
\centering
\setlength\tabcolsep{3pt}
\renewcommand{\arraystretch}{1.0}
\begin{tabular}{lccc}
\toprule
\hline
Dataset & Precision & Recall & F1 \\
\midrule
\hline
TempQuestion (Single answer) & $0.262$ & $ 0.262$ & $0.262$ \\
TempQuestion (Multiple answer) & $0.402$ & $0.155$ & $0.213$ \\
\textit{TempQA-WD} (Single answer) & $0.233$ & $0.233$ & $0.233$\\
\textit{TempQA-WD} (Multiple answer) & $0.271$ & $0.098$ & $0.140$\\
\hline
\bottomrule
\end{tabular}
\vspace{-0.1cm}
\caption{RAG-Token performance comparison on single answer questions vs. multiple answers questions.} 
\label{tab:rag_results}
\end{small}
\end{table}

\paragraph{Ablation Study of Only-Text Baseline:}
Table~\ref{tab:rag_results} shows the performance of RAG-Token model on questions with a single answer and multiple answers separately. As RAG is designed for factoid questions with a single answer, its recall is low in all the cases in Table~\ref{tab:rag_results}. In the case of multiple answers also RAG's performance is low compared to our proposed system
\paragraph{Extraction Evaluation:}
Extraction dataset is generated from TempQA-WD dev set. For each question in dev set, KB-specific lambda is decomposed into main lambda and auxiliary lambda. Main lambda is further converted into equivalent SPARQL to obtain the main candidates  and their intervals from the KB. Each of the main candidate is subsituted  into the main lambda to obtain equivalent NL query which is  also verified manually. The NL query along with the intervals for each main candidate become our extraction evaluation dataset.  
\fi
\section{Conclusion}
In this paper, we proposed an approach to combine the knowledge resources of KB (structured) and text (unstructured) for temporal QA. 
We used targeted extraction of temporal facts to compensate for KBQA failures.
The results of experimental evaluation show the effectiveness of our approach, usefulness of textual resources in helping KBQA. Future work includes improving extraction pipeline and extending to other types of reasoning.


\bibliography{anthology,custom}
\bibliographystyle{acl_natbib}

\end{document}